\pdfoutput=1

\documentclass[11pt]{article}

\usepackage{EMNLP2023}
\usepackage{graphicx}
\usepackage{times}
\usepackage{latexsym}

\usepackage[T1]{fontenc}

\usepackage[utf8]{inputenc}

\usepackage{microtype}

\usepackage{inconsolata}

\title{Inclusion of Role into Named Entity Recognition and Ranking}

\author{Neelesh Kumar Shukla \\
  Dept. of Computer Science \& Engg \\
  IIT Guwahati \\
  Assam, India\\
  {\tt neelesh.shukla@iitg.ac.in} \\\And
  Sanasam Ranbir Singh \\
  Dept. of Computer Science \& Engg \\
  IIT Guwahati \\
  Assam, India\\
  {\tt ranbir@iitg.ac.in} \\}

\begin{document}
\maketitle
\begin{abstract}
  Most of the Natural Language Processing systems are involved in entity-based processing for several tasks like Information Extraction, Question-Answering, Text-Summarization and so on. A new challenge comes when entities play roles according to their act or attributes in certain context. \textit{Entity Role Detection} is the task of assigning such roles to the entities. Usually real-world entities are of types: person, location and organization etc. Roles could be considered as domain-dependent subtypes of these types. In the cases, where retrieving a subset of entities based on their roles is needed, poses the problem of defining the role and entities having those roles. This paper presents the study of study of solving Entity Role Detection problem by modeling it as Named Entity Recognition (NER) and Entity Retrieval/Ranking task. In NER, these roles could be considered as mutually exclusive classes and standard NER methods like sequence tagging could be used. For Entity Retrieval, Roles could be formulated as \textit{Query} and entities as \textit{Collection} on which the query needs to be executed. The aspect of Entity Retrieval task, which is different than document retrieval task is that the entities and roles against which they need to be retrieved are indirectly described. We have formulated automated ways of learning representative words and phrases and building representations of roles and entities using them. We have also explored different contexts like sentence and document. Since the roles depend upon context, so it is not always possible to have large domain-specific dataset or knowledge bases for learning purposes, so we have tried to exploit the information from small dataset in domain-agnostic way.
\end{abstract}

\section{Introduction}
\label{section:introduction}
Consider a user reading news who comes across the following news article about a terrorist attack:

\begin{center}
\textit{The Anti-Terrorist Squad (ATS) of the U.P. police arrested an Indian Mujahideen operative and accused in the Delhi serial bomb blasts, Salman, in Siddhartnagar district on Friday night. A team of Delhi police had reached Lucknow to secure his transit remand for further investigations. Mr. Lal said Salman , who belongs to Sanjarpur village in Azamgarh district , was enrolled in the Bachelor of Computer Applications course on the Lucknow campus of Sikkim University, Manipal, in 2008.
}
\end{center} 

The user is interested in identifying organization which is behind the blasts or the location where blast has happened. In above example, multiple organizations have been mentioned like \textit{ATS}, \textit{Indian Mujahideen}, \textit{Delhi police} and \textit{Sikkim University} in which \textit{Indian Mujahideen} is the \textit{accused} where as \textit{Delhi} is the location of this blast not the other locations \textit{Siddhartnagar} or \textit{Lucknow}.

This is a use case of specific domains, where users are looking for entities with a specific purpose or act. This purpose or act could be defined as Role and the problem of assigning these roles to entities could be called Entity Role Detection. This paper proposes methods to automatically identify roles of entities.

We define the term \textit{Role} as the subtype of a named entity type that describes an entity by its act or association in a domain-specific corpus in a given context.

We study the problem of Entity Role Detection by modeling it as Named Entity Recognition (NER) and Entity Retrieval task. The roles are dependent upon the domain under consideration and many domains may not have large datasets or Knowledge Bases for learning. So, the approaches which make use of any kind of external resources haven't been explored and relied upon the knowledge gained from the corpus only.

The major contribution of this paper are as follows:
\begin{enumerate}
\item Introduces the idea of modeling Entity Role Detection task as NER and Entity Retrieval task.
\item Automated methods for representing Entity and Role (as information need) for entity retrieval.
\item Generic methods which could be applied to different domains or similar datasets.
\item Modeling relation between different mentions of an entity.
\end{enumerate}

\section{Motivation}

\label{section:motivation}

Several studies have been done on Entity Typing but a few have addressed \textit{Role}. The problem of identifying Roles is different than types in following sense:
\begin{itemize}
\item Roles are subtypes of coarse-grained types like person, location and organization which are specific to the event, topic or context.
\item Different domains may have different set of roles with varying semantics. With domains, comes the inherent issues of unavailability of domain-specific large dataset or Knowledge Bases (KBs) and need of domain-independent methods.
\item An entity may have multiple roles in the same document/event.
\item Roles of entity mentions depends upon other mentions of the same entity. e.g. In text excerpt \textit{ULFA was involved in attack happened in upper Assam. ULFA has set up a camp near Dibrugarh},  second mention of ULFA refers to first mention of ULFA for role.
\end{itemize}
Named Entity Recognition (NER) is one of the most important tasks in Information Extraction (IE) processes that consists in finding and classifying real-world entities denoted by a referent term or proper name (named entity). The current NER systems just deal with taxonomic types and are not meant to represent the entity's role in a specific context. 

Apart from above most of the NER methods rely upon entity's local context e.g. sentence level context using a window of nearby words which poses the following problem. Here the entities are identified and classified with an assumption that entities follow the similar sentence structure/sequence. In the problem of Role Detection such contexts may not be sufficient. For example, consider the news article example in the \ref{section:introduction} where first sentence could identify \textit{Salman} as accused  using the context which has information about being arrested and accusation. Whereas another mention of \textit{Salman} in sentence number three, may not able to identify its as accused using the sentence level context.

Fine-grained entity extraction and some of the Entity Retrieval tasks depends upon external knowledge sources like Wikipedia pages or Knowledge based for Entity identification and typing. Sometimes it is not possible to have or build such external knowledge bases and the only information possible is to get from corpus only.

Entity Retrieval tasks has focused on retrieving entities that satisfy a topic described in natural text or similar to the entities provided. A similar task of Question-Answering which ranks answers against questions, needs questions and answers in natural text. What if those descriptions are not known in advance? We would like to address this issue by learning such texts for Roles (information need or query).

Application-wise this Entity Role Detection problem could help in many applications like:
\begin{enumerate}
\item Event Summarization could be better not by giving details about involved entities as well as their role.
\item The entities which are playing a role, could be considered as prominent entity. Identifying similar entities with the similar roles could help in duplicate event detection.
\item Help in answering role-specific search queries like 'Who is victim in this event?'
\end{enumerate}

\section{Related Works}
Our work draws several related works from the task of Named Entity Recognition (NER), Fine-Grained Entity Recognition (FigER), Entity Retrieval, List Completion and Question-Answering task.

Named Entity Recognition is widely used task in Natural Language Processing (NLP). The task of NER is as formally defined by Message Understanding Conference 7 (MUC7) \cite{overview_muc7} considers name of people, location, organization, date time and currency entities. NER task naturally divided into two parts: Recognition: Finding entity names, and Classification: assigning appropriate semantic category. Early entity recognition systems adopted the rule-based approach which requires significant expert knowledge. There are various learning methods. Prominent methods are based on Decision Trees \cite{Isozaki2001}, Hidden Markov Model (HMM) \cite{Bikel1999}, Maximum Entropy Markov Models (MEMM) \cite{Chieu2003}, Support Vector Machines (SVM) \cite{Isozaki2002} and Conditional Random Field \cite{Duan2011}. All these methods rely on handcrafted features and researchers explored deep learning based methods \cite{Lample2016,Huang2015} using LSTM and BLSTM which has outperformed state-of-the-art named entity recognition systems. In Entity Role Detection problem the roles could be modeled as type classes , so that these methods could be used to solve this problem. The difference lies in the semantics of Role and Types defined in standard NER tasks. Types like person, location etc. which are used in NER are more syntactic and unrelated to topics. Whereas Roles exhibit latent semantic multi-facet relations. These relations are between topic and entity, between entities, even between roles itself as well as between mentions of the same entities. A recent work \cite{pcalleja-ranlp17:01} introduces Role Based Named Entity Recognition, which defines role as the patterns of word combinations. This needs domain knowledge as well as manual efforts to craft these patters. We have introduced models which are fully automatic and domain-agnostic.

Another dimension is Fine-Grained Named Entity Recognition, where the task is to classify an entity in broad set of types. Typically these systems use over hundred of labels and different mentions of the same entity has different labels. There have been many systems like \cite{Ling2012,Yosef2012,Gillick2014b,Ren2016,CoType2017}, proposed for Fine-grained task. These methods generally use external sources like Knowledge bases (KB) for entity typing. We have kept any external information source out of our scope and tried to utilize the information from dataset only.

Our work studies retrieval of entities against the roles as information need. There has been work in not just retrieving web pages or documents but also the objects like entities, books, properties etc. In 2005 TREC \cite{TREC2005} has introduced expert finding task where system needs to find knowledgeable people knowing about the topic provided. In 2007 List completion and Entity Ranking tasks has been introduced in INEX 2007 \cite{INEX2007} where they had a topic statements, example entity descriptions mostly using entity description from Wikipedia. Works \cite{Balog2009,Kaptein2013} have been done on ranking mechanism and describing entities, but no work describes role. Usually, Retrieval or Q\&A tasks have words, keywords or descriptions.But what if such descriptions and words are not known before hand. In our work, we have focused on identifying words/phrase so that both Entity and Role. A recent close work is \cite{Sarwar2018} which takes user's feedback to identify important terms. Whereas we are focusing on automatic approaches of identifying those terms.

The problem statement for Semantic Role Labeling is similar to our problem. Two systems Propbank \cite{PropBank2005} and FrameNet \cite{FrameNet1998} has been developed for this purpose. The problem with these systems is that they use thematic abstract roles, predefined rules or predicates and capture the sentence level syntax only. Different domain may have different general roles which could not be captured by such methods.

\section{Approach}
\label{section:approach}

The objective of our work is to study Entity Role Detection problem as Named Entity Recognition (NER) and Named Entity Retrieval task.

\subsection{Named Entity Recognition}
This section introduces Entity Role Detection problem as NER task where the objective is to classify entities which are person, location and organization into subtypes based on the role they play in the event.

Prominent Sequence labeling methods of NER, are supervised classification methods whose goal is to build a model whose input is a \textit{word sequence}, and output is a \textit{tag sequence}. Hidden Markov Model(HMM), Conditional Random Field(CRF) and Bidirectional Long Short Term Memory(BLSTM) based models were explored. We have used NLTK package \footnote{http://www.nltk.org/api/nltk.tag.html} for HMM and CRF tagging and the model defined in \cite{Lample2016} for BLSTM.

\subsection{Entity Ranking against Role}
An alternative approach to Entity Role Detection is Entity Ranking/Retrieval task. Information retrieval task needs the collection of certain resources and information need in form of query.  Entity Role Detection problem could be modeled as Entity Retrieval task by considering Role (or Type) as query and Entities present in a document or collections of documents as resources.  Our task is to rank these entities against the roles passed as query. 
This methods gives us the flexibility in two ways:
\begin{enumerate}
\item Dealing with a single role at a time.
\item Unlike sequence tagging problem, where all words needs to be correctly labeled with the tags (be it role tags or other tags) to get the valid sequence, we could focus on assigning roles to entities in which we are interested and try to rank them higher.
\end{enumerate}

Mostly entity retrieval works have focused on building representations of entities with very less focus on representing types or roles.

The main research questions in entity retrieval, similar to other retrieval tasks, can be organized around three main themes: 
\begin{enumerate}
\item How to represent entities (Entity Representation)? 
\item How to represent information needs (Type/Role Representation)? 
\item How to match these representations (Ranking Mechanism)?
\end{enumerate}

Similar to other retrieval method, we would like to represent the Query(Role) and Document(Entity) in form of natural text or a set of words or keywords.  As per our hypothesis, there are few words, which have the latent relation with the role. If somehow entities and role could be represented using those words, that could be one key step. In general, our approach would be to use these words to get the representation of entities and roles. 

\subsubsection{Word Representation/Embeddings}
The words need to be converted to feature vectors to be used for processing. Feature  engineering is a laborious task. Deep learning methods have been used in the industry and research community for their ability to learn optimal features automatically.  We will use Word2Vec \cite{Mikolov2013} based word representations for our study.  These word representations could also be used to achieve following:

We will also use these word representations to achieve following:
\begin{enumerate}
\item Word representation could be trained on corpus so that words could have corpus specific contexts.
\item We want to utilize word similarity in  entity as well as type contexts i.e nearby words.
\item Skip-gram model could be used/extended to represent Entity and Type or Role.
\end{enumerate}

\subsubsection{Entity Representation}
\label{subsection:entity_rep}
We are building entity representation on different granularity level context as defined below:
\subsubsection*{Entity Representation based on Sentence-Level Context}
\textit{Nearby Words as Context of Entity}:  An important observation is that nearby words play an important part in identifying role of an entity. Both Word2Vec and Glove word representation models have also utilized this relation of nearby words. We will define a context window of \textbf{previous d} and \textbf{next d} words, total window of size \textit{2d + number of words in entity}. After identifying these words we need to combine those words to represent entity. We are following three ways to do that:

\begin{enumerate}
\item \textit{Entity Word Vectors as Cluster (E-W-Nd)}: Use the set of embeddings of those nearby words as cluster representing entity.
\item \textit{Centroid Method (E-V-C-Nd)}: In this method average of embeddings of these nearby words are taken to represent the entity\cite{Wieting2015} also showed that a simple averaging over embeddings of the words is an effective representation.
\item \textit{Doc2Vec Method (E-V-C-Nd)}: Entity could of represented as the document containing those nearby words, we could use model presented in \cite{Le2014} which generates low dimensional vector representation of documents.
\end{enumerate}
\subsubsection*{Entity Representation based on Document-Level Context}
The idea is to use document level context to have better representation. In this work multiple mentions of the same entity has been utilized for this purpose. There could be multiple mentions ($m_1, m_2,...$) of an entity E in the document. While building representation for any entity mention $m_i$, we have considered the contexts of previous mentions of the same entity E, based on following assumptions.

\begin{enumerate}
\item \textbf{Majority Assumption:} At a document level, an entity may have multiple roles, but majority of the mentions of the same entity will have the same role,  which we call as the \textit{majority role} for that entity
\item \textbf{Positional Assumption:} The mention having majority role will appear before any other mention of the same entity.
\item \textbf{Information Flow Assumption:} The role information for an entity flows in top-down fashion, i.e.  the mentions which appear below in a document are influenced by mentions above them.
\end{enumerate}

\textbf{Out of Scope:}
In this work we are focusing on entity typing only not entity identification or entity  boundary identification. 

\subsubsection{Type(Query) Representation}
\label{subsection:type_rep}
Types or Roles are the information need or query, based on which the entities will be retrieved from the document.
\subsubsection*{Learning Type Representation using Skip-Gram model}
Similar to the idea of word representation learning presented in Skip-gram model, we tried to learn the representation of types by replacing entity mentions in the training set with their respective roles. This way we got a corpus having roles as a word itself. We learned representations of Role/Type using Skip-gram model, initializing weights for Role/Type words at random. For other words, we initialized their weights that we learned in Word Representation learning step.

We have two flavors of type representation with us:
\begin{enumerate}
\item \textit{Type Vector (TV)}: This is vector representation for role or type, learned as described above.
\item \textit{Query Expansion using Similar Word to Type Vector (TV-SW)}: Similar to idea of Query Expansion, we wanted to have more words in query. As per the idea presented in word representation learning methods, the words that appear in similar contexts will have the same vector representation as their neighboring words will be more or less same. We included those top-N words who have similar vector representations to that respective roles. We used cosine similarity measure here to find most similar words.
\end{enumerate}

\subsubsection{Word and Phrases}
While learning entity or role representations, we have analyzed token as single word (unigram) and phrases (bigram).

Phrase based approach has three steps as described below:

\subsection*{Step 1: Identifying Valid Phrases}
We have used following two approaches for identifying valid bigram phrases.
\begin{enumerate}
\item Collocations: Collocations are the words that appear frequently together and infrequently in other contexts.  Collocation based phrases were found based on the score described in \cite{Mikolov2013}
\item Relation  Phrases: Usually roles are some form of relations among entities or entity and event and so on.  These relation phrase could be used as a context for building representations.  To build such phrases, we have used Stanford Open Information Extraction \cite{Angeli2015} model which extracts relation tuples.
\end{enumerate}
\subsection*{Step 2: Learn Representation for Phrases}
Using the idea presented in Word2vec \cite{word2vecPaper}, these phrases could be replaced with tokens in the corpus to learn the representations using word representation learning models.
\subsection*{Step 3: Learn Representation for Entity and Role}
By replacing phrases by single token in the corpus, approaches described in sections \ref{subsection:entity_rep} and \ref{subsection:type_rep} could be used for the same.
\subsubsection{Ranking Mechanism}
We compute the similarity between entities present in the document with the given role(query) and rank these entities against that role based on the similarity score. 

\[SIM-GA(e,t) =
\frac{(\sum_{\vec{v_m}\epsilon e \cup t }\vec{v_m}^2) - (N_e + N_t)}{(N_e+N_t)(N_e+N_t-1)}
\]

where, $\vec{v}$ is the length-normalized vector and $N_e$ and $N_t$ are the number of vectors in e and t, respectively.

\begin{enumerate}
\item \textit{Representing Entity and Type as cluster of words vectors and computing Group-average similarity}: These two sets could be treated as cluster of vectors. We are using \textit{Group-average agglomerative clustering} which computes average similarity \textit{SIM-GA} of all pair of vectors including pairs from the same cluster. But self-similarities are not included in the average.
\item
\textit{Representing Entity and Type as single representation vector and Computing cosine similarity}: If Entity is represented as single vector and Type as another single vector. We could use SIM-GA considering each set containing one element only. We could use \textit{Cosine-Similarity}, both will give the similar results.
\end{enumerate}

\section{Experimental Setup}
\subsection{Dataset}
\label{subsection:dataset}
Dataset used in this paper is in-house built labeled dataset which consists of news articles about different bomb blast which has happened mainly between the year 2010 to 2014. These article were collected from the website of The Hindu, a leading media house in India. We have total 1037 articles. For the purpose of our work, entities in these articles were manually tagged with 10 different types (roles). These types and their brief description with examples are given below

\begin{itemize}
\item 
\textit{PER\_Victim}: Person who has suffered some kind of loss in the event of bomb-blast e.g. Congress leader \textbf{Mahendra Kumar} was killed.
\item
\textit{PER\_Accused}: Person who is responsible for the blast e.g. \textbf{Hafiz Sayeed} is alleged suspect in 28/11 Mumbai attacks.
\item
\textit{PER\_Others}: Other persons mentioned in the article e.g. Union minister \textbf{Shivraj Patil} today refrained from commenting on his offer of providing aids to victims.
\item
\textit{ORG\_Victim}:  Organization which has suffered some kind of loss in the event of bomb-blast e.g. \textbf{Congress} convoy was ambushed yesterday.
\item
\textit{ORG\_Accused}: Organization which is responsible for the blast e.g. \textbf{ISIS} claimed the responsibility of Paris attacks.
\item
\textit{ORG\_Others}: Other organizations mentioned in the article e.g. \textbf{United Nations} condemned Paris attacks.
\item
\textit{LOC\_Event}: Location where the blast/attack has occurred e.g. United Nations condemned \textbf{Paris} attacks.
\item
\textit{LOC\_Accused}: Location associated to accused persons/organizations. It includes the locations where they came from, stayed, home or where they got arrested, organization base, training camp location etc. e.g. NIA claimed that Dawood has given the instruction of attacks from \textbf{Karachi}.
\item
\textit{LOC\_Victim}: Locations associated with the victim person/organizations e.g. Those people who got injured in the attacks are on a visit from \textbf{United States}, \textbf{India} and \textbf{Japan}. 
\item
\textit{LOC\_Others}: Other location mentions in the article e.g President returned to \textbf{New Delhi} after visiting the victim of Mumbai blasts.
\end{itemize}

These 10 roles with their respective frequencies are displayed in table   \ref{tab:role_freq_tab}. We are dropping the Roles \textit{\_\_Others} and \textit{LOC\_Victim} for our study. \textit{\_\_Others} is dropped because entities not assigned any role will by default come under \textit{Others} and due to higher frequency of such roles, we were not able to see significant changes in the result while studying different approaches. The role \textit{LOC\_Victim} is dropped due to its very low frequency.

\begin{table}
\begin{center}
\begin{tabular}{|c|c|}
\hline
\textbf{Entity Role} & \textbf{Frequency}\\
\hline
PER\_Others & 3579\\
\hline
PER\_Victim & 506\\
\hline
PER\_Accused & 1132\\
\hline
ORG\_Victim & 227\\
\hline
ORG\_Accused & 1049\\
\hline
ORG\_Others & 3441\\
\hline
LOC\_Event & 3358\\
\hline
LOC\_Others & 3734\\
\hline
LOC\_Accused & 410\\
\hline
LOC\_Victim & 93\\
\hline
\end{tabular}
\end{center}
\caption{Roles and their frequencies}\label{tab:role_freq_tab}
\end{table}

\begin{table*}[t]
\centering
\begin{tabular}{|c|c|c|c|c|}
\hline
\textbf{Entity Role} & \textbf{HMM} & \textbf{CRF} & \textbf{BLSTM}\\
\hline
PER\_Victim & 61\% & 38\% &  57\% \\
\hline
PER\_Accused & 77\% & 63\%  & 58\% \\
\hline
ORG\_Victim & 24\% & 42\% &  40\% \\
\hline
ORG\_Accused & 86\% & 82\% & 79\% \\
\hline
LOC\_Accused & 22\% & 39\% & 58\% \\
\hline
LOC\_Event& 59\% & 57\% & 66\% \\
\hline
\hline
\textbf{Average Precision} & 54.96\% & 53.47\% & 59.75\%  \\
\hline
\end{tabular}
\caption{Role-wise Precision form HMM, CRF, BLSTM}
\label{tab:ner_seq_results}
\end{table*}

\begin{table*}[t]
\centering
\begin{tabular}{|c|c|c|c|c|c|c|}
\hline
\textbf{Method} & \textbf{mAP@1} & \textbf{mAP@2} & \textbf{mAP@3} & \textbf{mAP@4} & \textbf{mAP@5}\\
\hline
E-V-C-N10 & \textbf{55.29}\% & \textbf{62.13}\% & \textbf{64.44}\% & \textbf{65.24}\% & \textbf{65.29}\% \\
\hline
E-W-N10 & 53.20\% & 59.92\% & 62.98\% & 63.71\% & 63.96\% \\
\hline
E-V-D2V-N10 & 50.68\% & 58.28\% & 61.15\% & 62.55\% & 62.66\% \\
\hline
\end{tabular}
\caption{Performance of Various Methods of Entity Representation using Sentence Level Context in Entity Ranking against Roles Approach}\label{tab:entity_ranking_entity_rep_results}
\end{table*}

\begin{table*}[t]
\centering
\begin{tabular}{|c|c|c|c|c|c|c|}
\hline
\textbf{Method} & \textbf{mAP@1} & \textbf{mAP@2} & \textbf{mAP@3} & \textbf{mAP@4} & \textbf{mAP@5} \\
\hline
TV & \textbf{55.29}\% & \textbf{62.13}\% & \textbf{64.44}\% & \textbf{65.24}\% & \textbf{65.29}\% \\
\hline
TV-SW20 & 53.57\% & 61.39\% & 63.86\% & 64.19\% & 64.5\% \\
\hline
\end{tabular}
\caption{Performance of Various Methods of Type Representation using Sentence-Level Context in Entity Ranking against Roles Approach}\label{tab:entity_ranking_type_rep_results}
\end{table*}

\subsection{Experimental Setup}
As a basic preprocessing step, we have removed stopwords and performed stemming and lemmatization.

Since our dataset is very small, for effective learning, we trained word vector using Word2Vec Skip-gram model by initializing weights of the word from the publicly available 300-dimensional Glove word embeddings \footnote{https://nlp.stanford.edu/projects/glove/}.

For CRF, Word features are:\\

1.  Lower Case:  Non-capitalized words, bell.

2.  First 3 letters of word

3.  All caps:  Like name of organization, ISRO

4.  First word is Capital or rest in lower case:  Mr, Name 

For contextual features we used the word features of previous and next word also.\\

In Entity  Retrieval task,\\

1. Sentence context window size of -5 and +5 words/phrases for learning word, entity and role representations.

2. 20 most similar words/phrases are chosen to represent queries for types (roles).

3. We have removed duplicate entities from result while evaluating ranking performance. Assumption was that it will give space to more entities in top K position.

\subsection{Evaluation Metrics}
\begin{figure}[!h]
\centering
\includegraphics[scale=0.5]{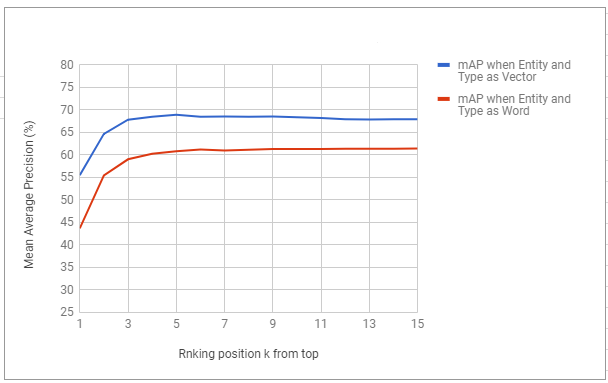}
\caption{Variation of Mean Average Precision w.r.t the Ranking Position K}
\label{fig:variation_map_wrt_k}
\end{figure}

We used \textit{Average Precision} to measure the performance of Named Entity Recognition's classification models. \textit{Mean Average Precision for top K positions (mAP@K)} metrics has been used to measure performance of Entity Ranking approch. As seen from figure \ref{fig:variation_map_wrt_k} the performance saturates after K=5. So, we have calculated mAP@1, mAP@2, mAP@3, mAP@4 and mAP@5 only.

\section{Results and Discussion}
\subsection{Named Entity Recognition}
From NER methods, we achieved average precision of 54.96\%, 53.47\% and 59.75\% with HMM, CRF and BLSTM respectively. Table \ref{tab:ner_seq_results} shows the role-wise performance of methods.

Usually such models give high performance of 90\% on CoNLL dataset \cite{conll-2003-task}. One of the key reason for bad performance is obviously the size of our dataset. Statistically it can be seen that, the entities in which are interested like Person Victim, Location Event are mentioned in around 5-10\% time only in an article, making them scarce to learn. Unavailability of large dataset is a big issue here.

These methods fail to capture the relation between multiple mentions of an entity as well as the relation between roles. A good observation is that sentence level context (sequence of surrounding words) influences the role to be assigned. With BLSTM it can be seen that low level word embeddings give better performance using deep learning techniques.

\subsection{Entity Ranking against Role}

\subsubsection{Experiments with different methods of entity representation} Experiments were performed with different methods of entity representation like centroid based (E-VC-N5), representative words (E-W-N5) and Doc2Vec (E-D2V-N5) method. We have kept Type representation (TV) same for all three to compare their performances. As per results in table \ref{tab:entity_ranking_entity_rep_results}, Centroid methods have performed best among all three.

\subsubsection{Experiments with different methods of type representation}
Experiments were performed to compare different methods of type (role) representation, type vector (TV) and representative top 20 similar words to the type vector (T-W-SW20) method. We have kept Entity representation as centroid (E-VC-N5) same. As per results in table \ref{tab:entity_ranking_type_rep_results}, TV has slightly performed better.

\subsubsection{Analysis of Different Representations}
In our study, We have seen vector representations performed well but representative words were not able to perform better. We could relate it to information task where document and queries are simply collection words and they need sophisticated methods to rank documents (a collection of words) against queries (another collection of word). But these representation should not be discarded because having these representative words for entity and type, helps us in using standard Information retrieval techniques of ranking like probabilistic information retrieval \cite{Petkova2007}, \cite{Robertson2009} , CNN based ranking \cite{Severyn2015} etc. In fact the vector representations are in way, a representation built over these words only.

\subsubsection{Experiment with Unigram and Bigram Phrases}
\begin{figure}[!h]
\centering
\includegraphics[scale=0.4]{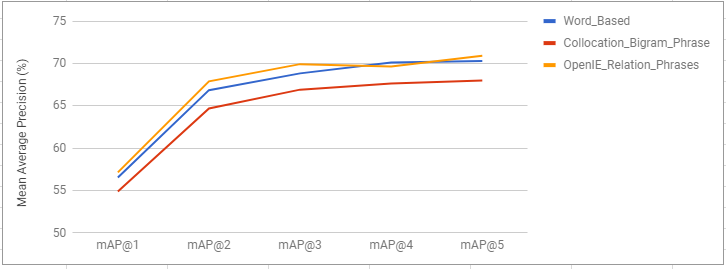}
\caption{Performance of Word vs Phrases Based Representation using Sentence-level Context}
\label{fig:word_vs_phrases}
\end{figure}

Experiments (figure \ref{fig:word_vs_phrases}) show that Relation-phrases have performed better than words but collocation based phrases have not performed well. The reason of bad performance of collocation is because of noisy phrases being added to the context like \textit{'on Sunday'}, \textit{'as part'}. More informative phrase like \textit{'blast at'}, \textit{'was attacked'} were not able to be formed in collocations. Whereas Relation phrase have captured such phrases which have latent relation to the role. There is improvement in mAP@1 and mAP@2 which shows that relevant entities have been pushed upwards using relation phrases.

\begin{figure}[!h]
\centering
\includegraphics[scale=0.4]{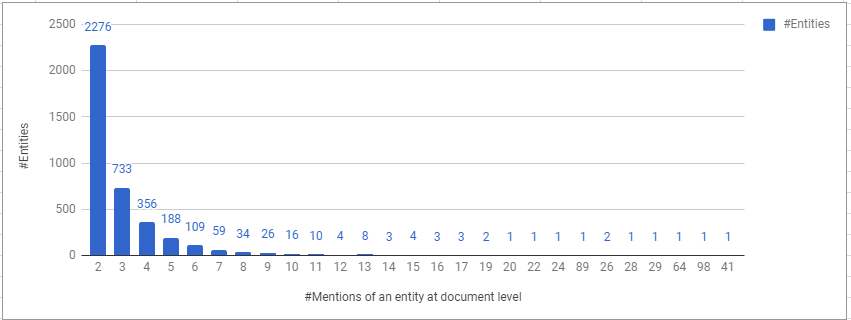}
\caption{Frequency Distribution of Entities having Multiple Mentions}
\label{fig:freq_dist_entity_wrt_num_mentions}
\end{figure}

\begin{figure}[!h]
\centering
\includegraphics[scale=0.4]{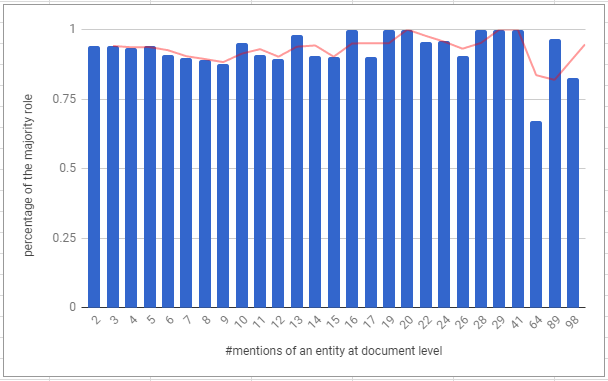}
\caption{Statistics for Majority Assumption: Percentage of Mentions having Majority Role}
\label{fig:majority_assumption_statistics}
\end{figure}

\begin{figure}[!h]
\centering
\includegraphics[scale=0.4]{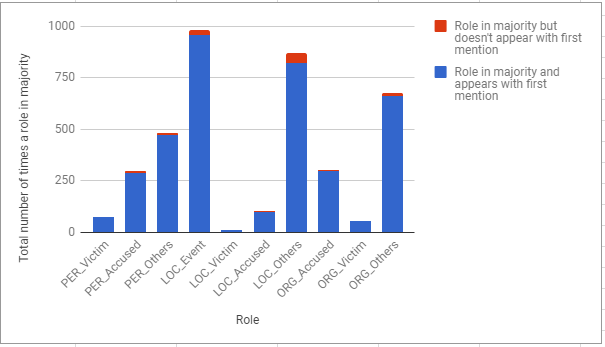}
\caption{Statistics for Positional Assumption}
\label{fig:positional_assumption_statistics}
\end{figure}

\begin{figure}[!h]
\centering
\includegraphics[scale=0.4]{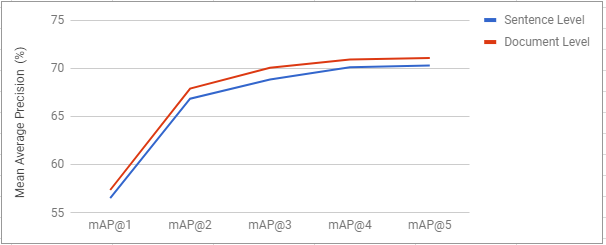}
\caption{Sentence vs Document Level Context using Word Based Approach, Centroid Representation of Entity and Vector Representation of Type (Role)}
\label{fig:sent_vs_doc_word_based}
\end{figure}

\begin{figure}[!h]
\centering
\includegraphics[scale=0.4]{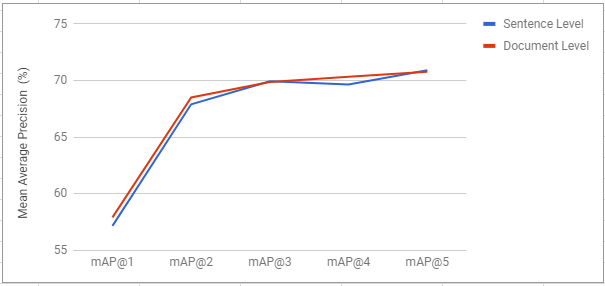}
\caption{Sentence vs Document Level Context using Relation Phrases Based Approach, Centroid Representation of Entity and Vector Representation of Type (Role)}
\label{fig:sent_vs_doc_rel_phrase_based}
\end{figure}

\subsubsection{Experiment with Sentence vs Document-level Context} 
Multiple mentions of an entity (at document level) has been considered to form document level context for an entity. In our dataset 23\% entities have multiple mentions. The distributions of mentions could be seen in figure \ref{fig:freq_dist_entity_wrt_num_mentions} e.g. 188 entities have 5 mentions at document level. Dataset statistics support majority assumption  as well as positional assumptions, discussed in section \ref{subsection:entity_rep}. In the dataset, a very high percentage, (83.67\%) of roles, assigned to an entity at document level, are same. For example, from bar chart in figure \ref{fig:majority_assumption_statistics}, it can be seen that Entities having 10 mentions (at document level) will have average 95\% of the mentions with the same role. Also 98\% of the entity mentions having majority role, appear first before any other mention of the same entity. From the figure \ref{fig:positional_assumption_statistics}, it can be seen 97\% of time LOC Event is the majority role as well as first entity mention assigned role.
Having previous mentions context words into current mention (of the same entity) context has shown little improvement but not the significant improvement. The reason is that the contexts of lower level mentions have been polluted by intermediate mentions. For example consider following set of sentences having mentions of entity ULFA:

i) ULFA was involved in attack happened in upper Assam. 

ii) ULFA has set up a camp near Dibrugarh. 

iii) ULFA has sent the militants here or last Monday.

All these ULFA mentions are referring to Role \textit{ORG\_Accused}. The first mention could be assigned role from its sentence level context only as we have phrases like \textit{involved in} attack. But next two mentions need help from first mention. It can be seen that for third mention of ULFA, contexts have been polluted by the second mention of ULFA.

\section{Conclusion and Future Work}
In this paper, we study the problem of \textit{Entity Role Detection} by formulating it as Name Entity Recognition and Entity Retrieval task. We proposed automated methods to identify word and phrases which could represent entity and their respective roles.
We found out that standard sequence tagging methods of NER task like HMM, CRF and BLSTM are giving satisfactory results considering the size of the dataset. In Entity Retrieval, we showed that phrases especially relation phrases are better representative of roles than words. Also different entities (their mentions) and roles have a latent relation, which could be utilized to have better representation.
There is a significant room of improvement in identifying those representative words/phrases for the role as well as entities. 

As future work, Methods could be developed to identify informative words in the context and combine those words/phrases to have better representation of Entity and Role. Standard Ranking methods like probabilistic information retrieval or deep learning based Learning to Rank models could developed and used. 
Also, semantic relations between mentions, entities and roles could be modeled to have better approaches at different granularity levels \cite{Tatbul2018}.

\bibliography{anthology,custom}
\bibliographystyle{acl_natbib}

\end{document}